\pdfoutput=1

\documentclass[11pt]{article}
\usepackage{acl}

\usepackage{algorithm}
\usepackage{algpseudocode}
\usepackage{tikz}

\usepackage{booktabs} 
\usepackage{tabularx} 
\usepackage{geometry} 
\usepackage{multirow}
\usepackage{times}
\usepackage{latexsym}
\usepackage{xspace}
\usepackage{amsmath}
\usepackage{amsfonts}
\usepackage{changepage} 
\usepackage[most]{tcolorbox}  
\usepackage{listings}
\usepackage{float}

\usepackage[T1]{fontenc}

\usepackage[utf8]{inputenc}

\usepackage{microtype}

\usepackage{inconsolata}

\usepackage{graphicx}

\newcommand{\ourtool}{IPOMP\xspace}

\newcommand{\stagetwo}{IPOMP$_{Stage2}$\xspace}

\newcommand{\stageone}{IPOMP$_{Stage1}$\xspace}

\newcommand{\stageoneRandom}{IPOMP$_{Random}$\xspace}

\title{Model Performance-Guided Evaluation Data Selection for Effective Prompt Optimization}

\author{Ximing Dong\textsuperscript{1}, Shaowei Wang\textsuperscript{2}\thanks{\;\;Corresponding author.},  Dayi Lin\textsuperscript{1}, Ahmed E. Hassan\textsuperscript{3} \\
\textsuperscript{1}Centre for Software Excellence, Huawei, Canada \\
\textsuperscript{2} Department of Computer Science, University of Manitoba, Canada \\
\textsuperscript{3}School of Computing, Queen's University, Canada\\
\texttt{\{ximing.dong,dayi.lin\}@huawei.com, shaowei.wang@umanitoba.ca, ahmed@cs.queensu.ca}\\}

\begin{document}


\maketitle
\begin{abstract}

Optimizing Large Language Model (LLM) performance requires well-crafted prompts, but manual prompt engineering is labor-intensive and often ineffective. Automated prompt optimization techniques address this challenge but the majority of them rely on randomly selected evaluation subsets, which fail to represent the full dataset, leading to unreliable evaluations and suboptimal prompts. Existing coreset selection methods, designed for LLM benchmarking, are unsuitable for prompt optimization due to challenges in clustering similar samples, high data collection costs, and the unavailability of performance data for new or private datasets. To overcome these issues, we propose \ourtool, an \textbf{I}terative evaluation data selection for effective \textbf{P}rompt \textbf{O}ptimization using real-time \textbf{M}odel \textbf{P}erformance. \ourtool is a two-stage approach that selects representative and diverse samples using semantic clustering and boundary analysis, followed by iterative refinement with real-time model performance data to replace redundant samples. Evaluations on two datasets BIG-bench and LIAR, and two models GPT-3.5 and GPT-4o-mini, show that \ourtool improves effectiveness by at least 1.6\% to 3.1\%, and stability by at least 50\% to 55.5\% compared with the best baseline across the studied datasets and models, with minimal computational overhead below 1\%. Furthermore, the results demonstrate that our real-time performance-guided refinement approach can be universally applied to enhance existing coreset selection methods.

\end{abstract}

\section{Introduction}\label{sec:intro}

Given a task, drafting an effective prompt is a key part of optimizing the Large Language Model’s (LLM) performance~\citep{kojima2022large,pryzant2023automatic,wei2022chain}. Minor changes in prompt can lead to performance gains or losses, necessitating
prompt engineering for effective LLM utilization~\citep{liu2023pre}. 
To avoid the process of manually creating prompts, recent work aims to automate the process of generating natural language prompts that are also interpretable~\citep{APEzhou2022large,zhang-etal-2023-auto,EVOPROMPTguo2023connecting,OPROyang2024largelanguagemodelsoptimizers,zhang2022tempera,deng2022rlprompt}. 

All the proposed approaches require optimizing prompts by evaluating them over an evaluation dataset. However, using the entire training dataset is impractical and cost-prohibitive~\citep{pacchiardi2024100,albalak2024survey}. Consequently, most approaches randomly select a small subset of samples from the entire training data to evaluate new prompts~\citep{APEzhou2022large,zhang-etal-2023-auto,EVOPROMPTguo2023connecting,OPROyang2024largelanguagemodelsoptimizers,pryzant2023automatic}. However, random selection often fails to produce representative samples of the entire training dataset, leading to unreliable evaluation results~\citep{zadrozny2004learning} and under-optimized prompts. No existing approaches have been proposed to select representative samples for evaluating prompts for prompt optimization. 

Various coreset selection approaches have been developed for benchmarking machine learning models, with the core idea being to select representative samples that effectively represent the entire dataset based on different criteria such as semantics~\citep{sener2017active,har2004coresets}, model performance indicators such as confidence scores~\citep{pacchiardi2024100,vivek2023anchor,polo2024tinybenchmarks}, and training errors~\citep{paul2021deep}. However, these approaches are not well-suited for prompt optimization. In semantics-based approaches, samples for certain tasks (e.g.,  Navigation task from BIG-bench~\citep{srivastava2023beyond}) tend to be highly similar, making it challenging to cluster them effectively solely based on their semantics. On the other hand, approaches that leverage model performance information rely on evaluation results from previously tested models to predict the performance of new models~\citep{pacchiardi2024100,vivek2023anchor,zhou2023predictable,zhou2022reject}. However, such approaches have notable limitations. Firstly, model performance data for training samples is not always available, particularly for new or proprietary datasets. Even if it is possible, collecting the model performance information prior to the prompt optimization process is expensive. Secondly, using past model performance to estimate the capabilities of current LLMs often results in sub-optimal predictions, as model behaviors may vary significantly (which is evidenced by our results in Section~\ref{sec:rq1}).

To address the limitations of existing coreset selection methods and tailor them for prompt optimization, we propose a two-stage approach called \ourtool that leverages both semantic and model performance information. In the first stage, we identify informative samples by clustering the entire training dataset based on semantic similarity and selecting representative samples from each cluster. Additionally, to enhance diversity, we incorporate boundary cases by selecting the most distant sample pairs in the semantic space. In the second stage, we iteratively refine the evaluation samples by incorporating their real-time model performance during the optimization process. Specifically, we identify redundant samples based on their performance across the generated prompts and replace them with contrasting samples.  

We evaluated IPOMP on the BIG-Bench and LIAR datasets, comparing its performance against several SOTA baseline methods. \ourtool outperformed all baselines, achieving effectiveness improvements in terms of Accuracy ranging from 1.6\% to 3.1\% and significantly enhancing stability by at least 50\% in terms of standard deviation, while with a computational overhead of less than 1\%. Furthermore, our evaluation results demonstrate that our real-time model performance-guided refinement approach in the second stage can be universally adapted to existing coreset selection approaches to enhance their effectiveness and stability.

\section{Background and related work}\label{sec:background}

\subsection{Prompt Optimization}
Prompt optimization refers to the systematic process of designing, refining, and evaluating prompts to improve the performance of large language models (LLMs) on specific tasks~\citep{APEzhou2022large}. 
Let $\mathcal{L}$ denote a large language model, $T$ represent a given task, and $D^{\text{evaluation}} = \{x_i, y_i\}_{i=1}^N$ be the evaluation dataset, where $x_i$ are inputs and $y_i$ are the corresponding desired outputs for $T$. A prompt $P$ is a sequence of tokens that guides $\mathcal{L}$ to generate outputs $\hat{y}_i = \mathcal{L}(x_i, P)$. 

The objective of prompt optimization is to find a prompt  $P^*$ that maximizes the task performance over $D^{\text{evaluation}}$. This can be expressed as:  

\[
P^* = \arg\max_{P} \mathcal{M}\left(\{\mathcal{L}(x_i, P) y_i\}_{i=1}^N\right),
\]  

where $\mathcal{M}$ is a performance metric that quantifies the alignment between the model-generated outputs \( \hat{y}_i = \mathcal{L}(x_i, P) \) and the ground-truth outputs \( y_i \), such as Accuracy, F1-score, BLEU score, or task-specific measures.

The optimization typically involves iterative refinement of prompt $P$ using various strategies, which typically can be categorized into two families: \textit{non-directional} and \textit{directional}. Non-directional approaches sample or generate new inputs randomly and do not explicitly aim to reduce error on a train set over the optimization iteration based on feedback, such as random search~\citep{APEzhou2022large,zhang-etal-2023-auto} and evolutionary algorithm~\citep{EVOPROMPTguo2023connecting,OPROyang2024largelanguagemodelsoptimizers,fernando2309promptbreeder}. For instance, APE generates semantically similar candidate prompts for a task based on their performance on a training subset and iteratively selects the best prompt\citep{APEzhou2022large}. In \textit{Directional} family, the generation of new prompts is guided by the error measure on evaluation data, such as using gradient~\citep{pryzant2023automatic,juneja2024task} and reinforcement learning~\citep{zhang2022tempera,deng2022rlprompt,yao2023retroformer}. For instance, APO uses minibatches of data to form natural language gradients that criticize the current prompt~\citep{pryzant2023automatic} . The gradients are then propagated into the prompt by editing the prompt in the opposite semantic direction of the gradient. 

Most existing prompt optimization approaches either utilize the entire training dataset or randomly sample a subset, which can make the evaluation process overly expensive or suboptimal. To address this challenge, we propose a novel evaluation data selection approach specifically designed for prompt optimization.


\subsection{Coreset selection for benchmarking machine learning models}

We are the first to propose evaluation data selection in the context of prompt optimization. The most related field to our work is coreset selection. Coreset selection aims to find the most informative subset $D^{\text{Core}}$ $\subset$ $D^{\text{Training}}$ with the constraint $|D^{\text{Core}}|$ $\ll$ $|D^{\text{Training}}|$, so that the model trained on $D^{\text{Core}}$ has close generalization performance to the model trained on the whole training set $D^{\text{Training}}$.

Numerous approaches have been developed for coreset selection for evaluating machine learning models in recent years. Geometry-based methods assume that semantically similar data points share properties~\citep{chen2012super,sener2017active,sinha2020small,agarwal2020contextual}. However, solely relying on semantics while overlooking the model performance could lead to sub-optimal performance in identifying the representative samples, typically for tasks where the samples are naturally semantically close to each other. To improve accuracy, performance-based approaches consider factors such as confidence~\citep{coleman2019selection,margatina2021active,lin2023optimal,kim2020confident}, error~\citep{paul2021deep,toneva2018empirical,liu2021just}. For instance, approaches based on confidence prioritize uncertain samples, assuming they have a greater impact on model performance. Error-based approaches assume that training samples are more important if they contribute more to the error or loss when training models. 
Decision boundary-based approaches focus on samples near the decision boundary, as they are harder to classify and valuable for coreset selection~\citep{ducoffe2018adversarial,margatina2021active,chai2023efficient}. More recently, methods have leveraged evaluation results from previously tested LLMs to predict the performance of new models~\citep{pacchiardi2024100,vivek2023anchor,zhou2023predictable,zhou2022reject}.

We propose a novel two-stage data selection approach for prompt optimization that leverages both semantic and real-time model performance features. Unlike existing model performance-based methods that require a preliminary stage to gather performance data or rely on prior model results to predict new outcomes~\cite{vivek2023anchor,pacchiardi2024100,zhou2022reject}, our approach dynamically collects performance data during optimization in real-time, ensuring greater accuracy and cost-efficiency. Additionally, it can be seamlessly integrated into any prompt optimization method involving iterative refinement.



\section{Methodology}\label{sec:method}


We propose a two-stage approach that leverages both semantic and model performance features to select evaluation data for effective prompt optimization: 1) Diverse sample selection; and 2) Real-time model performance-guided iterative refinement. In stage 1, we cluster training samples based on semantics and select representative samples from each cluster, while incorporating boundary cases by selecting the most distant pairs in the semantic space. In stage 2, we iteratively refine the selected samples by analyzing real-time model performance during optimization, removing redundant samples, and replacing them with contrasting ones. For simplicity, we use the terms ``coreset selection'' and ``data selection'' interchangeably in the following text.


\subsection{Stage 1: Diverse sample selection}\label{sec:stage1}

\algnewcommand\algorithmicinput{\textbf{Input:}}
\algnewcommand\algorithmicoutput{\textbf{Output:}}
\algnewcommand\Inputs{\item[\algorithmicinput]}
\algnewcommand\Outputs{\item[\algorithmicoutput]}

\begin{algorithm}
    \caption{Diverse sample selection}
    \label{alg:stage1}
    \begin{algorithmic}[1]
        \Inputs  Training set $D^{\text{training}}$, number of selected samples $N$ , number of clustering groups $k$, portion of samples from semantic clustering $\alpha$
        \Outputs $D^{evaluation}$ of size $N$
         \State \textcolor{blue}{\# Select $\alpha N$ samples based on semantic clustering}
        \State $S_{\text{clustering}}  \leftarrow \{\}$
        \State $clusters \gets \operatorname{KMeans}(D^{\text{training}}, k)$
        \State $S_{\text{clustering}} \gets \text{sampleProportionly}(\alpha N, clusters)$
        \State $D^{\text{training}}.\text{remove}(S_{\text{clustering}})$
        \State \textcolor{blue}{\# Select $(1-\alpha) N$ boundary samples}
        \State $n, S_{\text{boundary}} \leftarrow 0, \{\}$
        \While{$n \leq \alpha N$}
            \State $d_{1}, d_{2} \leftarrow \text{getLeastSimilarPair}(D^{\text{training}})$
            \If{$d_1$ $\notin$ $S_{\text{clustering}}$}
                \State $S_{\text{boundary}}.\text{add}(d_1, d_2)$
                \State $n \leftarrow n + 2$
            \EndIf
            \State $D^{\text{training}}.\text{remove}(d_1, d_2)$
        \EndWhile
        \State Return  $D^{\text{evaluation}} \gets S_{\text{clustering}} + S_{\text{boundary}}$
        
\end{algorithmic}
\end{algorithm}

In this stage, we aim to select a small subset from the entire training set that comprehensively represents all training samples. One common strategy is to first cluster samples of the training set based on their selected properties, such as semantic~\citep{sener2017active,har2004coresets}. Then it samples representatives from each resultant cluster to form a reduced set. However, such clustering-based approaches probably would miss boundary cases~\citep{huang2024semantic}. 
Therefore, in this stage, we combine semantic clustering and boundary selection methods to select a small yet comprehensive subset of samples from the training set. 

Given the training set $D^{\text{training}}$ which contains \( M \) training samples  \(\{D_1, D_2, \dots, D_M\}\), our algorithm outputs a subset $D^{\text{evaluation}}$ with a size of \( N \), where $N$ $\ll$ $M$. We demonstrate the algorithm in Algorithm~\ref{alg:stage1}, which consists of two steps: 
1) Selecting informative samples using semantic clustering. We first embed each sample in $D^{\text{training}}$ into latent space. We utilize Sentence-Bert~\citep{reimers2019sentence} to encode semantic representations. We then use K-means to cluster samples into $k$ clusters (Line 3). Note that K-means is selected due to its effectiveness and efficiency. We do not use methods to determine the value of $k$ since they typically require additional time. In addition, throughout experiments, we find that the value of $k$ does not impact our data selection approach significantly (see more results in Appendix~\ref{sec:rq5}). Lastly, we randomly select samples proportionally from each cluster based on their size and totally select $\alpha N$ samples (i.e., $S_{clustering}$), where $\alpha$ is the portion of selected samples from semantic clustering (Line 4).
2) Identifying boundary cases. Inspired by prior study~\citep{huang2024semantic}, we select boundary cases by finding samples that are least similar to each other. To do so, similar to step 1, we embed the samples into semantic latent space and find the pairs of samples having the furthest distance, iteratively (Lines 8 - 14). Note that we only include the samples that are not included in $S_{\text{clustering}}$.
Calculating the distance among all samples is expensive, with a time complexity of $O(d N^2)$, where $d$ is the dimension of embeddings and $N$ is the size of dataset. To improve the efficiency, we first detect the boundary points in the latent space by following previous study~\cite{angiulli2002fast}, and then find the furthest pairs among those boundary points. 
Finally, we combine the clustering samples $S_{\text{clustering}}$ and boundary samples $S_{\text{boundary}}$ together as the final subset of samples.


\begin{algorithm}
    \caption{Real-time model performance-guided iterative refinement.}
    \label{alg:stage2}
    \begin{algorithmic}[1]
        \Inputs  Selected samples from stage 1 $D^{\text{evaluation}}$, Replace rate $\beta$, Training set $D^{\text{training}}$, Correlation threshold $CT$, Prompt optimization approach $\mathcal{PO}$, LLM $\mathcal{L}$, number of iterations $I$
        \Outputs Refined samples $D_{\text{refined}}^{\text{evaluation}}$, best prompts $bestPrompt$
        \State $i, S_i \leftarrow 0, D^{\text{evaluation}}$
    \While{$i < I$}
        \State $candP \leftarrow \mathcal{PO}.\text{updatePrompts}(i, candP)$
        \State $MP_{\text{runtime}} \leftarrow  \text{recordPerf}(S_{i}, \mathcal{L}, candP)$
        \State \textcolor{blue}{\#Identify redundant samples in $S_{i}$ based on their model performance}
        \State $ clusters \leftarrow \text{Clustering}(MP_{\text{runtime}},CT)$
        \State $ S_{redundant} \leftarrow \text{sampleRedundant}(clusters,\beta)$
        \State \textcolor{blue}{\# Find least samples to replace}
        \For{$e_i \in S_{redundant}$} 
            \State $d \leftarrow \text{leastSimSample}(e_i, D^{\text{training}})$
            \State $S_{redundant}.\text{replace}(e_i,d)$
        \EndFor
    \EndWhile
    
    \State Return $D_{\text{refined}}^{\text{evaluation}} \leftarrow S_i$, $bestPrompt \leftarrow identifyBest(candP)$
\end{algorithmic}
\end{algorithm}

\subsection{Stage 2: Real-time model performance-guided iterative refinement}\label{sec:stage2}


We obtain a reduced set $D^{\text{evaluation}}$ based on their semantics after applying Algorithm~\ref{alg:stage1}. However, solely relying on semantics while overlooking the model performance could lead to sub-optimal performance, typically for tasks where the samples are naturally semantically close to each other. To address this, we design a novel algorithm called Real-time model performance-guided iterative refinement, which updates $D^\text{evaluation}$ iteratively by replacing redundant samples with contrasting ones based on their model performance. This process leverages real-time model performance observed during prompt optimization, eliminating the need for pre-collected performance data from existing models or preliminary evaluations.

\begin{figure}[h]
    \centering
    \vspace{-0.3in}
    \includegraphics[width=0.45\textwidth]{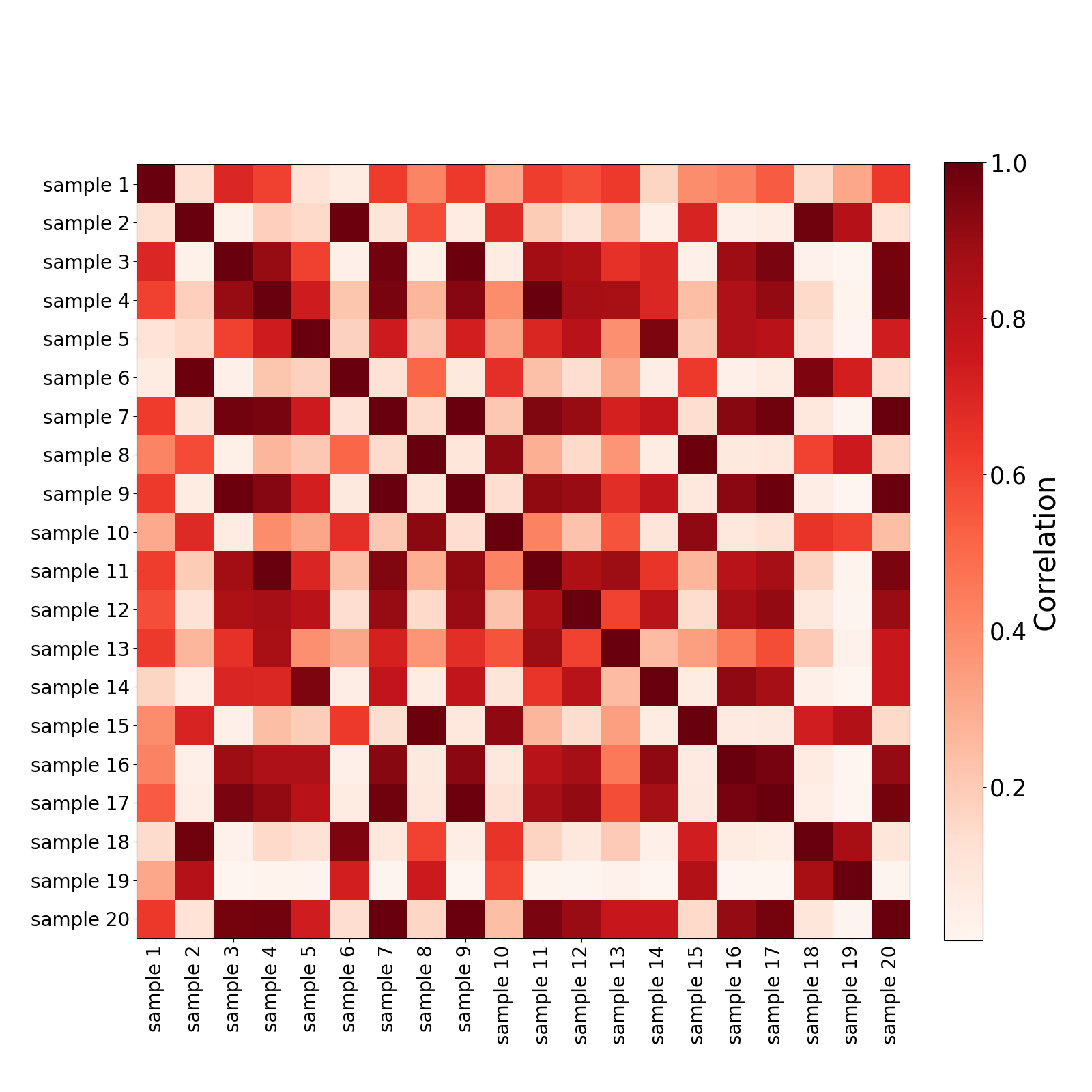}
    \caption{Correlation among the samples selected in stage 1 based on their real-time performance across candidate prompts during the initial iteration of APE. Each cell represents the correlation between a pair of samples.}
    \vspace{-0.1in}
    \label{fig:sm}
\end{figure}

Our approach is inspired by a key observation that a significant portion of samples exhibit high correlations in their model performance across prompts during prompt optimization. Figure~\ref{fig:sm} presents a heatmap where each cell represents the correlation between two samples based on their real-time model performance (i.e., logits in this case). As we can see, a substantial portion of samples (20\% in this case) exhibit a correlation greater than 0.9 with others. This indicates redundancy among these samples and they can be replaced with alternative samples from the training set to enhance the diversity of $D^\text{evaluation}$. 

We demonstrate our algorithm in Algorithm~\ref{alg:stage2}. Given a prompt optimization technique $\mathcal{PO}$, for each iteration, our algorithm first identifies the redundant samples in the $D^\text{evaluation}$ (Lines 3 - 7) and replaces a certain portion of them with the opposite (i.e., the most dissimilar) ones retrieved from the training set. 
To be specific, for each iteration, we record the performance of each example's performance across candidate prompts on the LLM $\mathcal{L}$ (line 4). The performance matrix $MP_{runtime} \in \mathbb{R}^{|S_i| \times (|output| \times |candP|)}$ where $|S_i|$ is the number of samples in $S_i$, $|output|$ is the size of output labels, and $|candP|$ is the number of candidate prompts $candP$ generated by prompt optimization approach $\mathcal{PO}$ in each iteration. We use logits as the proxy of the model's confidence to construct the performance matrix. For instance, if the output labels of a task are True/False, the matrix includes two dimensions to represent the performance: $\{Logit(True), Logit(False)\}$. If the output is True, the performance is $\{Logit(True), 0\}$. In cases where the output is neither True nor False, the performance is set to $\{0, 0\}$. To identify the redundant samples, we use a hierarchical cluster to build clusters and group samples that share high correlations in the same cluster (Line 6) by following prior studies~\citep{wang2018understanding,rajbahadur2017impact}. In our case, we set the threshold $CT$ to 0.9 (i.e., highly correlated). We then randomly select a portion (i.e., $\beta$) of samples $S_{redundant}$ from each cluster with highly correlated samples and replace them. For replacement, we iteratively select samples from $D^\text{training}$ which have the lowest semantic similarity for examples in $S_{redundant}$ (Line 9 - 12). We assume that the pair with the lowest semantic similarity is more likely to yield answers, which leads to different model performances.
For efficient search, we use Hierarchical Navigable Small World (HNSW)~\citep{malkov2018efficient} to perform an approximate search by inverting its similarity function to calculate the dissimilarity between the query examples and examples in the training data. HNSW is efficient in high-dimension space.


\section{Experimental Setting}\label{lab:experimentalsetting}


\subsection{Datasets}
In this paper, we evaluated our approach on two datasets \textbf{BIG-bench} dataset \cite{srivastava2023beyond} and \textbf{LIAR}~\cite{wang2017liar}. We selected the following five tasks from Big-bench. \textbf{Presuppositions as Natural Language Inference (NLI)}, where to reason whether a presupposition is embedded in a given statement and output entailment, neutral, or contradiction. \textbf{Navigation}: Given a sequence of navigational instructions, the objective is to ascertain if an agent would return to the original point of departure. \textbf{Implicatures}: This task requires models to determine whether a response given by one speaker to another constitutes an affirmative or negative reply. \textbf{Metaphor Understanding}: This task presents a model with a metaphoric sentence and requires it to identify whether a subsequent sentence accurately interprets the initial metaphor. \textbf{Sports Understanding}: This task requires models to assess whether a synthetically constructed sentence related to a sport is plausible or not. 
We also evaluated \textbf{LIAR}~\cite{wang2017liar}, a widely used public dataset for \textbf{Fake News Detection}. 

We measure the \textbf{Accuracy} of those classification tasks to evaluate the effectiveness of the studied coreset selection approaches. In addition, we use the \textbf{standard deviation} (SD) to measure the stability of the approaches. 

\vspace{-1.5mm}

\subsection{Prompt optimization approaches and base LLMs}
As discussed in Section~\ref{sec:background}, prompt optimization can primarily be categorized into two families, directional and non-directional. We selected two state-of-the-art approaches APO ~\citep{pryzant2023automatic} and EVOPROMPT~\citep{EVOPROMPTguo2023connecting} from non-directional family, and one of the most commonly used approaches APE~\citep{APEzhou2022large} from directional family. We use their default setting in our experiments and choose GPT-3.5 and GPT-4o-mini. 

\subsection{Baselines}
To evaluate the effectiveness of our approach, we compare our proposed approach with various baselines. 
\textbf{Random.} Randomly sample examples from the training set. \textbf{Clustering.} Selecting examples proportionally from the clusters that are constructed based on samples' semantics as discussed in Section~\ref{sec:stage1}. \textbf{Boundary.} We select boundary cases as discussed in Section~\ref{sec:stage1} by following prior study~\citep{huang2024semantic}. Our work is the first to address evaluation data selection for prompt optimization, and no existing state-of-the-art (SOTA) approaches are available. Therefore, we benchmark our approach against two SOTA coreset selection methods designed for LLM evaluation. \textbf{Anchor-Point}~\citep{vivek2023anchor}.
This method clusters examples based on the model's confidence in the correct class and selects representative examples, called ``anchor point'' as the coreset. To adapt it for prompt optimization, we collect the model's confidence scores by running the training dataset through a set of 10 prompts generated via prompt optimization in a preliminary stage. \textbf{Prediction-based}~\citep{pacchiardi2024100}.
This method predicts an instance's performance on an LLM by training a generic assessor on existing LLM performance data. We adapt this approach by training the assessor on our dataset, BIG-bench, using GPT-3.5, following~\cite{pacchiardi2024100}. The trained assessor predicts performance on a set of initial prompts, and examples are clustered based on these predictions, similar to Anchor-Point. See more details in Appendix~\ref{sec:baselineImpl}. Note that we do not consider the entire training data as a baseline, since it is too expensive and infeasible in practice.

\subsection{Implementation Details}
In our experiments, the size of $D^{\text{evaluation}}$ is set to 20, and the number of clustering groups $K$ is set to five by default unless otherwise specified. For \ourtool, we set $\alpha$ to 0.5, meaning that half of the samples in $D^{\text{evaluation}}$ come from the boundary method, while the remaining half comes from the clustering method. The correlation threshold $CT$ and the replace rate $\beta$ are set to 0.9 and 0.5, respectively. 
We run each task of each method five times and report the average performance. 

\section{Results}\label{results}

\subsection{Effectiveness and stability}\label{sec:rq1}

\begin{table*}[]
\centering
\caption{Comparison of the effectiveness (Accuracy) and stability (SD) of the studied prompt optimization approaches with different evaluation data selection approaches.}
\vspace{-0.1in}
\renewcommand{\arraystretch}{1.2} 
\resizebox{2.1\columnwidth}{!}{
\begin{tabular}{l|cccccc|cccccc}

\hline
& \multicolumn{6}{|c|}{\textbf{GPT-3.5 - BIG-bench}} & \multicolumn{6}{c}{\textbf{GPT-4o-mini - BIG-bench}} \\
\hline
                  & \textbf{Random} & \textbf{Boundary} & \textbf{Clustering} & \textbf{Anchor-Point} & \textbf{Prediction-based} & \textbf{\ourtool}
                  & \textbf{Random} & \textbf{Boundary} & \textbf{Clustering} & \textbf{Anchor-Point} & \textbf{Prediction-based} & \textbf{\ourtool}\\ 
\hline
\textbf{EVOPROMPT }      & 0.743$\pm$0.028 & 0.757$\pm$0.022   & 0.759$\pm$0.031   & 0.774$\pm$0.028 & 0.758$\pm$0.025    & \textbf{0.776\textcolor{red}{$\uparrow$}$\pm$0.017\textcolor{green}{$\downarrow$}} 
& 0.694 $\pm$ 0.047 &	0.715 $\pm$ 0.042 &	0.706 $\pm$ 0.037 &	0.756 $\pm$ 0.026 &	0.709 $\pm$ 0.047  & \textbf{0.758\textcolor{red}{$\uparrow$}$\pm$0.011\textcolor{green}{$\downarrow$}}
\\ 
\textbf{APO} & 0.691$\pm$0.040 & 0.718$\pm$0.052   & 0.723$\pm$0.025   & 0.750$\pm$0.020 & 0.743$\pm$0.042    & \textbf{0.753\textcolor{red}{$\uparrow$}$\pm$0.009\textcolor{green}{$\downarrow$}} 
&0.703 $\pm$0.038 &	0.72$\pm$ 0.043 &	0.701 $\pm$ 0.022 &	0.743$\pm$ 0.032 &	0.690 $\pm$0.043 &	\textbf{0.780\textcolor{red}{$\uparrow$}$\pm$0.012\textcolor{green}{$\downarrow$}} \\ 
\textbf{APE}      & 0.722$\pm$0.037 & 0.708$\pm$0.045   & 0.684$\pm$0.032   & 0.727$\pm$0.035 & 0.707$\pm$0.048    & \textbf{0.742\textcolor{red}{$\uparrow$}$\pm$0.010\textcolor{green}{$\downarrow$}} 
& 0.717 $\pm$ 0.040	& 0.734 $\pm$ 0.030 &	0.770 $\pm$ 0.030 &	0.770 $\pm$ 0.023 &	0.716 $\pm$ 0.042 &	\textbf{0.794\textcolor{red}{$\uparrow$}$\pm$0.012\textcolor{green}{$\downarrow$}} \\ 

\textbf{Average}  & 0.719$\pm$0.035           & 0.727$\pm$0.040             & 0.725$\pm$0.029             & 0.745$\pm$0.028          & 0.725$\pm$0.038              
& \textbf{0.757\textcolor{red}{$\uparrow$}$\pm$0.012\textcolor{green}{$\downarrow$}} 
& 0.704$\pm$ 0.041 &	0.723 $\pm$ 0.038&	0.725 $\pm$ 0.029&	0.756 $\pm$ 0.027	& 0.705 $\pm$ 0.044	&  
\textbf{0.778\textcolor{red}{$\uparrow$}$\pm$0.011\textcolor{green}{$\downarrow$}} \\
\hline


& \multicolumn{6}{|c|}{\textbf{GPT-3.5 - LIAR}} & \multicolumn{6}{c}{\textbf{GPT-4o-mini - LIAR}} \\
\hline
                  & \textbf{Random} & \textbf{Boundary} & \textbf{Clustering} & \textbf{Anchor-Point} & \textbf{Prediction-based} & \textbf{\ourtool}
                  & \textbf{Random} & \textbf{Boundary} & \textbf{Clustering} & \textbf{Anchor-Point} & \textbf{Prediction-based} & \textbf{\ourtool}\\ 
\hline
\textbf{EVOPROMPT }    
& 0.753 $\pm$ 0.042 &	0.798 $\pm$ 0.037&	0.772 $\pm$ 0.035&	0.810 $\pm$ 0.030&	0.734 $\pm$ 0.043
& \textbf{0.818\textcolor{red}{$\uparrow$}$\pm$0.015\textcolor{green}{$\downarrow$}} 
& 0.756 $\pm$ 0.045 &	0.806 $\pm$ 0.041 &	0.782 $\pm$ 0.037	&0.816 $\pm$ 0.026	&0.754 $\pm$ 0.043&	 \textbf{0.838\textcolor{red}{$\uparrow$}$\pm$0.012\textcolor{green}{$\downarrow$}} 
\\ 
\textbf{APO} & 0.732 $\pm$ 0.039&	0.792 $\pm$0.032	& 0.731 $\pm$ 0.031	& 0.794 $\pm$ 0.028&	0.754 $\pm$ 0.038&	\textbf{0.812\textcolor{red}{$\uparrow$}$\pm$0.014\textcolor{green}{$\downarrow$}} 
&0.746 $\pm$ 0.059&	0.792 $\pm$ 0.061&	0.776 $\pm$ 0.037&	0.806 $\pm$ 0.022&	0.754 $\pm$ 0.048&	 \textbf{0.836\textcolor{red}{$\uparrow$}$\pm$0.013\textcolor{green}{$\downarrow$}} \\ 
\textbf{APE}   &   0.743 $\pm$ 0.043 &	0.721 $\pm$ 0.035&	0.753 $\pm$ 0.042&	0.801 $\pm$ 0.023&	0.748 $\pm$ 0.037	& \textbf{0.832\textcolor{red}{$\uparrow$}$\pm$0.011\textcolor{green}{$\downarrow$}} 
& 0.746 $\pm$ 0.04 &	0.792 $\pm$ 0.039&	0.808 $\pm$ 0.042&	0.8 $\pm$ 0.025	&0.742 $\pm$ 0.04&	 \textbf{0.826\textcolor{red}{$\uparrow$}$\pm$0.011\textcolor{green}{$\downarrow$}} \\ 

\textbf{Average}  & 0.742 $\pm$ 0.041&	0.770 $\pm$ 0.035	&0.752 $\pm$ 0.036&	0.801 $\pm$ 0.027&	0.746 $\pm$ 0.039  
& \textbf{0.820\textcolor{red}{$\uparrow$}$\pm$0.012\textcolor{green}{$\downarrow$}} 
& 0.748 $\pm$ 0.048	&0.797 $\pm$ 0.047&	0.788 $\pm$ 0.038&	0.807 $\pm$ 0.024&	0.75 $\pm$ 0.043&	 
\textbf{0.833\textcolor{red}{$\uparrow$}$\pm$0.012\textcolor{green}{$\downarrow$}} 
\\ \hline

\end{tabular}
}
\label{tab:effectiveness}
\vspace{-0.1in}
\end{table*}

\textbf{All coreset selection approaches enhance the effectiveness and stability of prompt optimization techniques compared to Random selection. Among these, \ourtool demonstrates superior performance.}
Table~\ref{tab:effectiveness} presents the effectiveness and stability of all studied coreset selection approaches, across different prompt optimization techniques. Comparing Random with all other data sampling approaches, including \ourtool, we observe a significantly superior performance across all prompt optimization techniques, which indicates that selecting representative samples is important for prompt optimization. \ourtool improves the best baseline (Anchor-Point) by at least 1.6\% to 3.1\% across the studied datasets and models. Typically for APE, the improvement gained from \ourtool is larger than other prompt optimization techniques, which is probably attributed to its nature, where in each iteration, the prompt is optimized based on the evaluation (e.g., gradient) from the last iteration, the evaluation data's quality is typically important for the success of the prompt optimization. On the other hand, we summarize the standard deviation (SD) across all datasets for each selected baseline. We find that \ourtool exhibits greater stability than other baselines, achieving the lowest standard deviation across prompt optimization techniques, with an improvement of at least 50\%. This is typically attributed to our real-time model performance-guided iterative refinement strategy (i.e., stage 2 of \ourtool), which dynamically refines the evaluation data during runtime and guarantees the stability of our approach. See more ablation analysis in Section~\ref{sec:rq2}.

Boundary and Clustering share similar effectiveness across all studied prompt optimization techniques. However, compared with \ourtool, it demonstrates lower effectiveness and suffers lower stability, which indicates that relying solely on semantics to obtain samples is insufficient, and incorporating model performance offers a promising approach to identifying representative examples. On the other hand, Prediction-based requires adaptation to new datasets and exhibits lower effectiveness compared to \ourtool, making it limited compatibility in the context of prompt optimization. 

Anchor-Point consistently ranks as the second-best approach across all prompt optimization methods. Notably, approaches utilizing real-time model performance data (\ourtool and Anchor-Point) outperform those relying solely on semantics or prior model data, as performance feedback offers more precise insights for distinguishing samples. \ourtool surpasses Anchor-Point by combining semantic and real-time performance data, meanwhile achieving better efficiency. Unlike Anchor-Point, which requires a costly preliminary stage to collect model confidence, \ourtool gathers performance data in real-time during the optimization process.


\subsection{Ablation Analysis}\label{sec:rq2}

\begin{table*}[htbp]
\centering
\caption{Comparison of the effectiveness and stability of the studied prompt optimization approaches with \ourtool and its variants.}
\scriptsize
\renewcommand{\arraystretch}{1.1} 
\resizebox{0.95\textwidth}{!}{%
\begin{tabular}{l|ccc|ccc}
\hline

& \multicolumn{3}{c|}{\textbf{GPT-3.5 - BIG-bench}} & \multicolumn{3}{c}{\textbf{GPT-4o-mini - BIG-bench}}\\
\hline

                  & \textbf{\stageone} & \textbf{\stageoneRandom} & \textbf{\ourtool} & \textbf{\stageone} & \textbf{\stageoneRandom} & \textbf{\ourtool}
                  \\ \hline

\textbf{EVOPROMPT}         & 0.745$\pm$0.029   & 0.751$\pm$0.021    & \textbf{0.776$\pm$0.017}    
& 0.737$\pm$0.014 &	0.754$\pm$0.012	& \textbf{0.758$\pm$0.011} \\

\textbf{APO}               & 0.730$\pm$0.031   & 0.738$\pm$0.015    & \textbf{0.753$\pm$0.009  }   
& 0.739$\pm$0.012	& 0.757$\pm$0.012 &	\textbf{0.780$\pm$0.012    }
\\ 

\textbf{APE}               & 0.724$\pm$0.042   & 0.723$\pm$0.027    & \textbf{0.742$\pm$0.010  }   
& 0.754$\pm$0.012	& 0.705$\pm$0.013 &	\textbf{0.794$\pm$0.012  }   \\ 

\textbf{Average}           & 0.733$\pm$0.034   & 0.737$\pm$0.021    & \textbf{0.757$\pm$0.012}        
& 0.743$\pm$0.012	& 0.738$\pm$0.012 &	\textbf{0.778$\pm$0.011}
\\ 
\hline


& \multicolumn{3}{c|}{\textbf{GPT-3.5 - LIAR}} & \multicolumn{3}{c}{\textbf{GPT-4o-mini - LIAR}}\\
\hline

                  & \textbf{\stageone} & \textbf{\stageoneRandom} & \textbf{\ourtool} & \textbf{\stageone} & \textbf{\stageoneRandom} & \textbf{\ourtool}
                  \\ \hline

\textbf{EVOPROMPT}         & 0.802$\pm$0.019&	0.807$\pm$0.017	&\textbf{0.818$\pm$0.015}	&0.820$\pm$0.015	&0.809$\pm$0.014 &	\textbf{0.838$\pm$0.011}\\

\textbf{APO}               & 0.801$\pm$0.021 &	0.812$\pm$0.018 &	\textbf{0.812$\pm$0.014} &	0.797$\pm$0.022 &	0.824$\pm$0.015 &	\textbf{0.836$\pm$0.011}
\\ 

\textbf{APE}               &0.812$\pm$0.018	&0.829$\pm$0.013&	\textbf{0.832$\pm$0.011} &	0.801$\pm$0.014&	0.826$\pm$0.015&	\textbf{0.827$\pm$0.013}  \\ 

\textbf{Average}           & 0.805$\pm$0.020 &	0.816$\pm$0.016	&\textbf{0.820$\pm$0.013}&	0.806$\pm$0.017	&0.820$\pm$0.015	& \textbf{0.833$\pm$0.012}
\\ 
\hline

\end{tabular}
}
\label{table:2}
\end{table*}

\noindent\textbf{\stageone vs. \ourtool} To evaluate the contribution of model performance-guided iterative refinement (i.e., \stagetwo), we conduct an ablation analysis. We compare \ourtool with its variant in which sage 2 is removed (i.e., \stageone). As shown in Table~\ref{table:2}, on average, without stage 2, the effectiveness of \ourtool drops 2.4\%. In addition, the performance becomes unstable without stage 2. Specifically, the standard deviation increases by a factor of 2.83, indicating that stage 2 significantly enhances stability. Actually stage 2 indeed reduce the redundant samples. For instance, in APO, after the first round of refinement of stage 2, the redundancy of examples (correlation $> 0.9$) are significantly reduced from 19\% to 10\% (see Appendix~\ref{sec:casestudy}).

\noindent\textbf{\stageoneRandom vs. \ourtool} To evaluate the importance of the diverse sample selection of \ourtool
(i.e., \stageone), we construct a variant, namely \stageoneRandom, where we replace stage 1 with random sampling and keep the rest of \ourtool unchanged. As shown in Table~\ref{table:2}, the accuracy of \stageoneRandom is substantially 2\% lower than \ourtool, illustrating the necessity of diverse data selection at the beginning. 
In summary, both stages of \ourtool make significant contributions to the effectiveness and stability of \ourtool.

\begin{figure*}[h]
    \centering
    \includegraphics[width=0.45\textwidth]{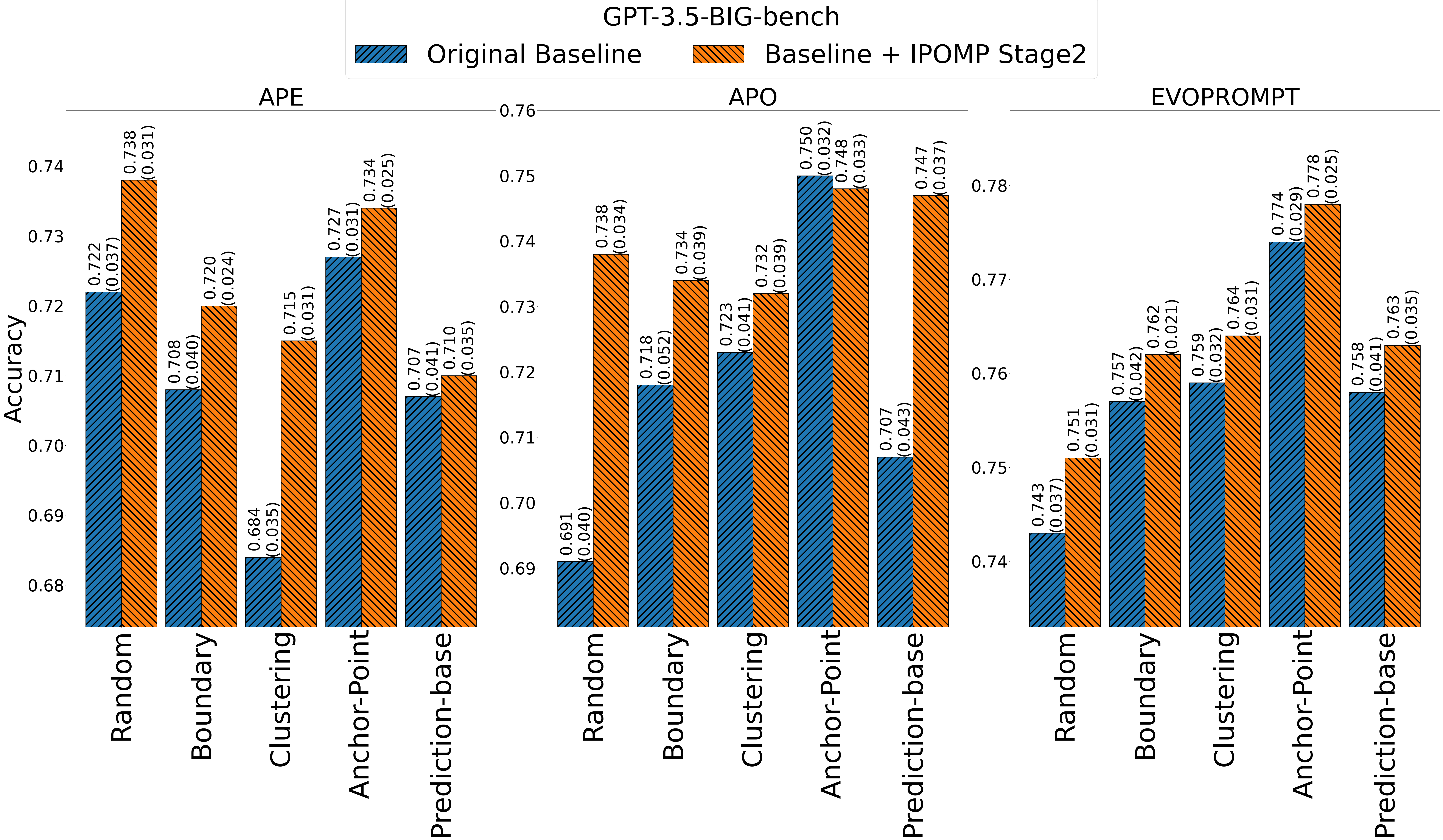}
    \includegraphics[width=0.45\textwidth]{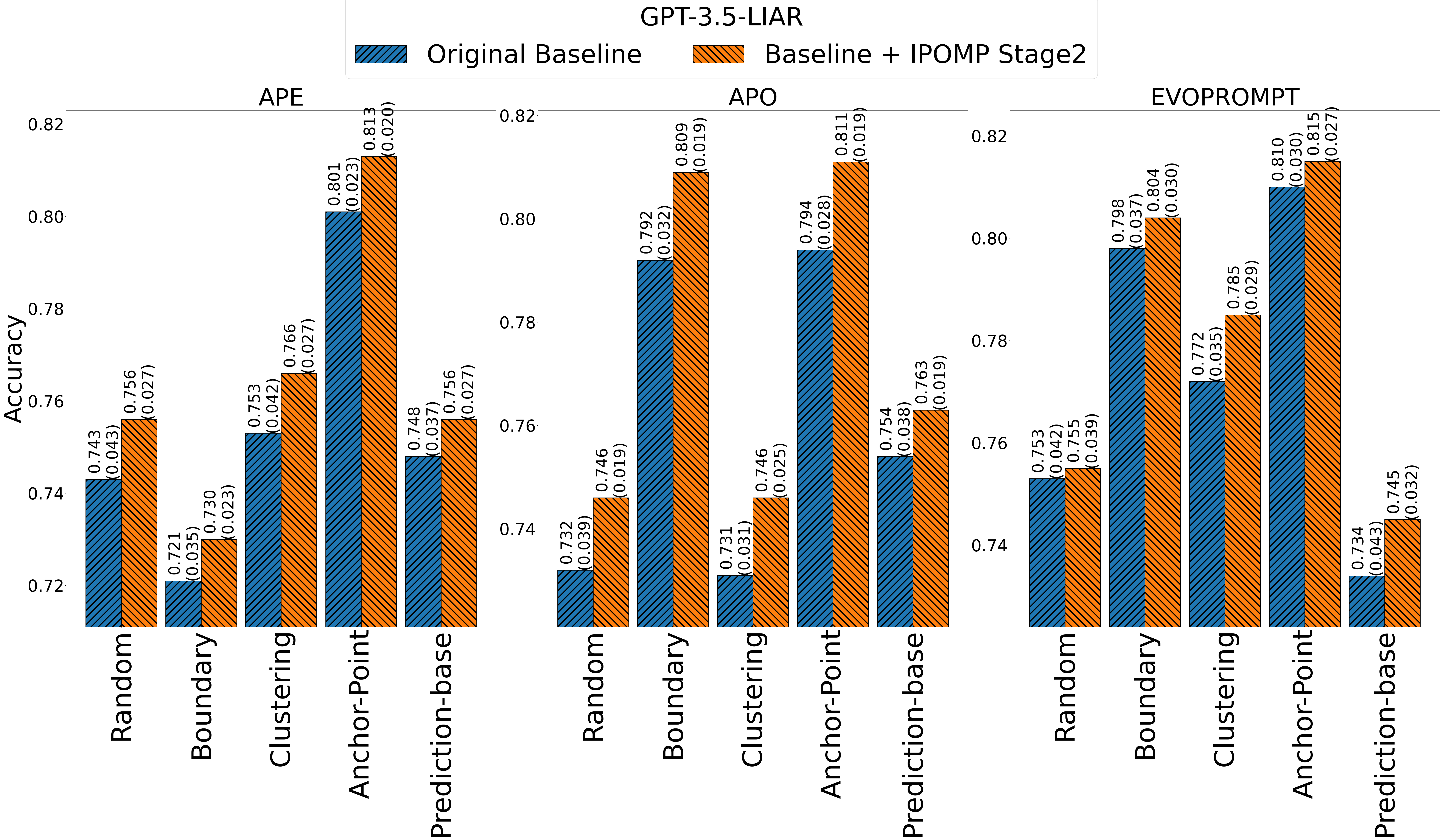}
     \includegraphics[width=0.45\textwidth]{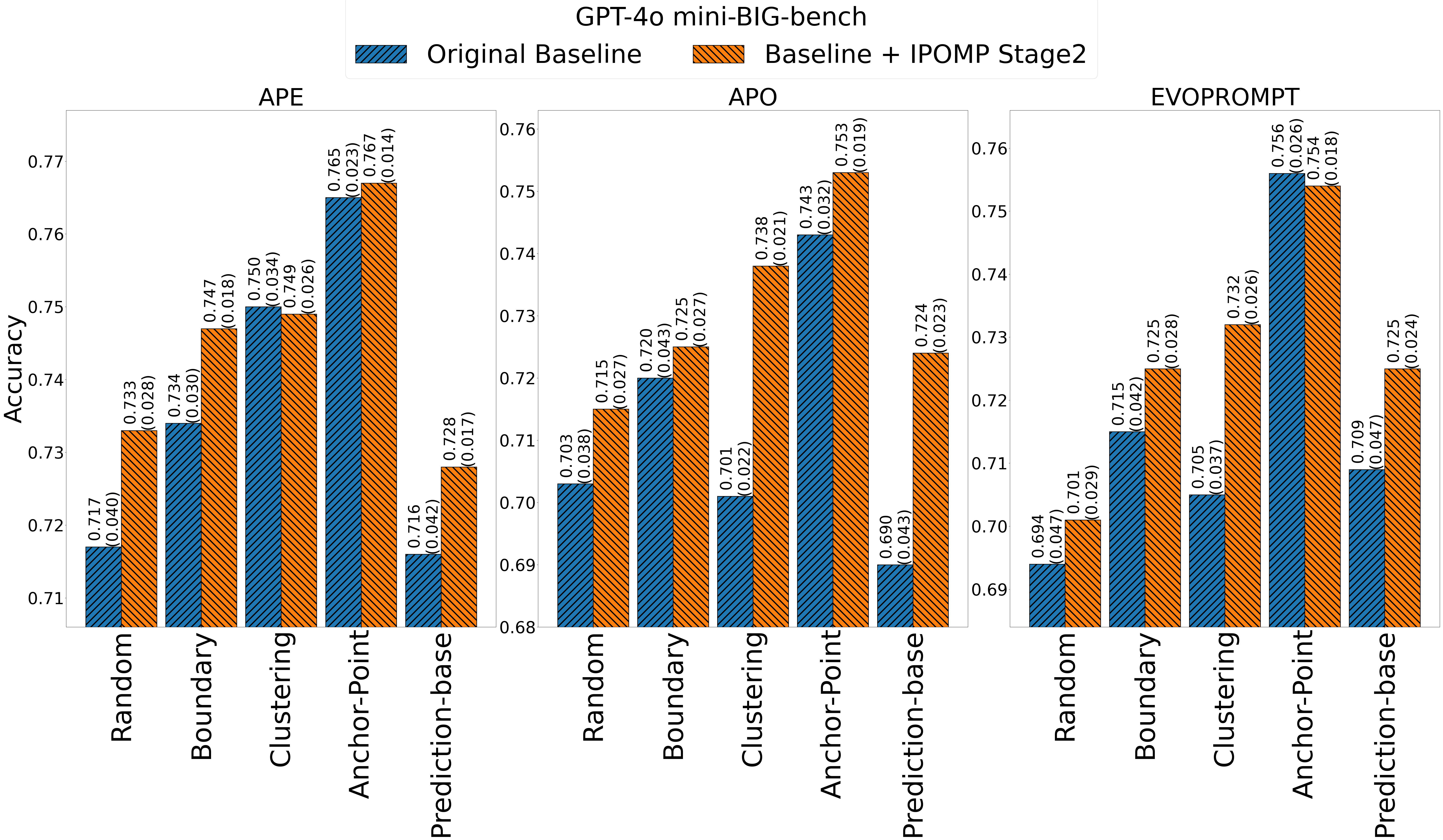}
      \includegraphics[width=0.45\textwidth]{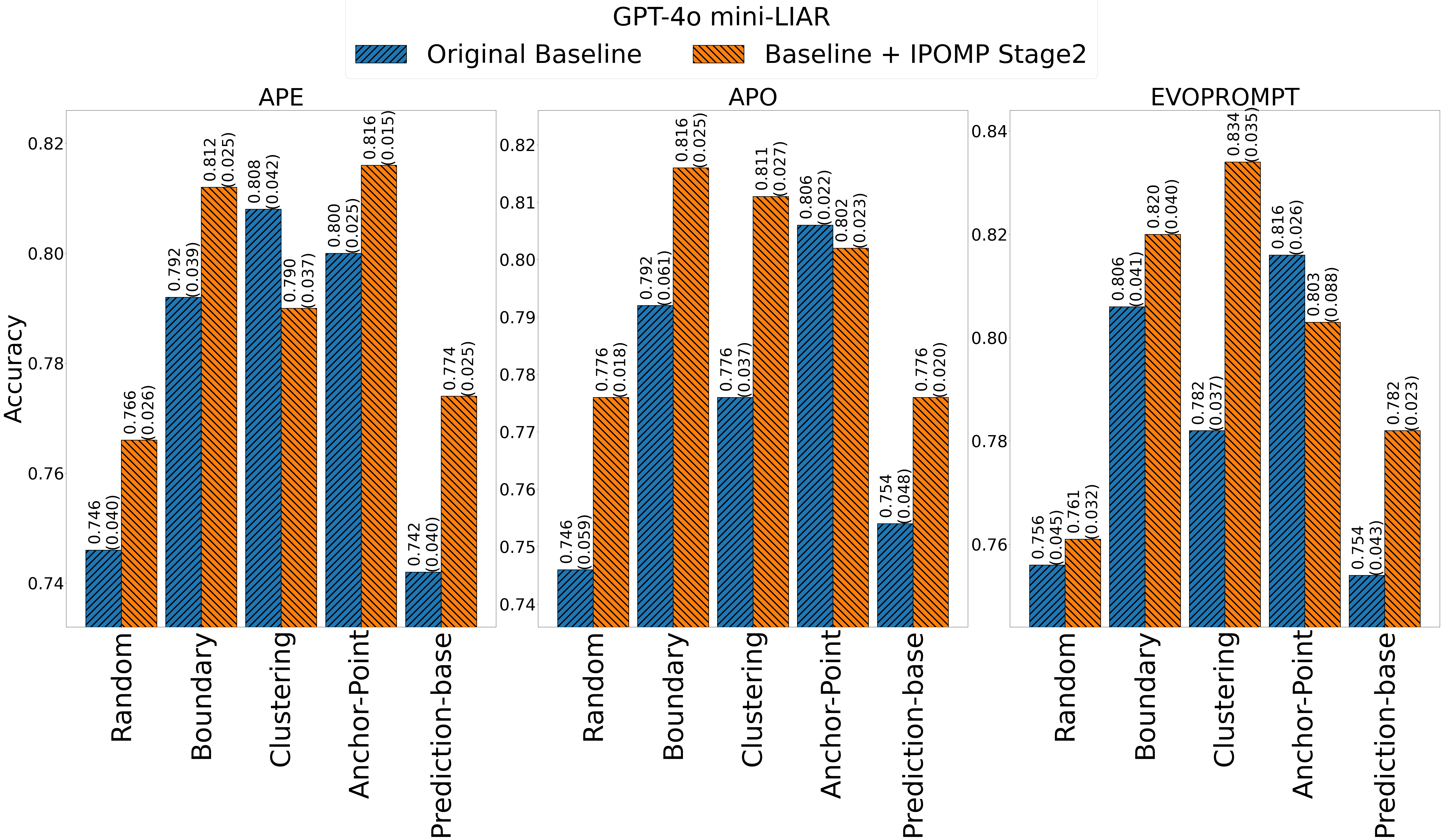}
    \caption{Comparison of the effectiveness of original baselines and Baseline+\stagetwo. SD is indicated in parentheses.}
    \vspace{-0.1in}
    \label{fig:importancestage2}
\end{figure*}







\begin{table*}[hbt!]
\centering
\caption{Average execution time (in seconds) of the studied prompt optimization after applying \ourtool (including time required for each stage) and other baselines on BIG-bench when using GPT-3.5. Note that Random can be considered as the \textbf{pure execution time} of prompt optimization techniques as the execution time of random sampling is negligible. For Anchor-Point, we also present the time for the preliminary stage to collect the model confidence information.}
\renewcommand{\arraystretch}{1.2} 
\resizebox{0.95\textwidth}{!}{%
\begin{tabular}{lcccccccc}
\hline

                  & \textbf{\stageone} & \textbf{\stagetwo} & \textbf{\ourtool}  & \textbf{Random} & \textbf{Boundary} & \textbf{Clustering} & \textbf{Anchor-Point} & \textbf{Prediction-based}
                  
                  \\ 
\hline
\textbf{APO}               &0.45   &2.74  &470.86      &469.74 &470.34 &470.83 &469.74+200.36 &480.57 \\
\textbf{APE}               &0.37   &2.31  &120.23       &109.84 &110.67 &111.12 &113.26+205.32 &123.31 \\ 
\textbf{EVOPROMPT}         &0.51   &3.43  &613.75      &609.35 &611.53 &608.38 &609.31+207.85 &622.61 \\ 
\textbf{Average}           &0.45   &2.83  &401.61       &396.31 &397.51 &396.61 &397.43+204.51 &405.50  \\ \hline\end{tabular}
}
\label{table:time}
\end{table*}

\noindent\textbf{Baseline vs. Baseline+\stagetwo} Besides enhancing \ourtool, our real-time model performance-guided iterative refinement can serve as a versatile plugin alongside any data sampling approach to refine evaluation data during runtime. To assess its effectiveness, we apply it to all selected baselines. Figure~\ref{fig:importancestage2} illustrates the performance of these baselines before and after incorporating \stagetwo. 
The performance of all baselines improves after applying our real-time refinement component (i.e., \stagetwo), except for Anchor-Point in some cases. For instance, on average, \stagetwo improves the performance of Random, Boundary, Clustering, Anchor-Point, and Prediction-based baselines by 2.3\%, 1.1\%, 1.5\%, 0.3\% and 1.6\% when using GPT-3.5 on BIG-bench dataset, respectively. More importantly, incorporating \stagetwo into the original baselines improves the stability significantly. For instance, \stagetwo reduces the standard deviation by 18.8\%, 60.0\%, 6.9\%, 10.8\%, and 16.8\% for Random, Boundary, Semantic, Anchor-Point, and Prediction-based when using GPT-3.5 on BIG-bench, respectively. In summary, \stagetwo not only boosts the performance of baselines but also enhances their stability. Our results demonstrate that \textbf{\stagetwo is a practical and adaptable enhancement for various data selection methods, effectively refining evaluation data by leveraging real-time model performance insights.}

\subsection{Impact of sample size}\label{sec:rq3}

\begin{figure}
    \centering
    \includegraphics[width=0.5\textwidth]{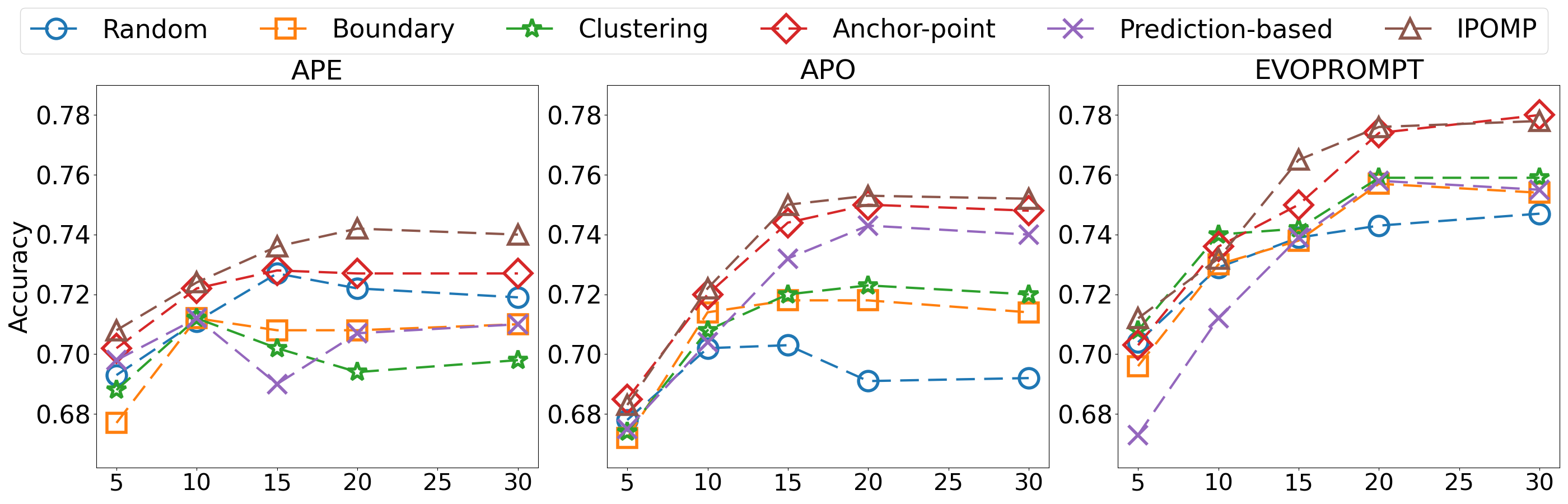}
    \caption{The impact of different sample sizes selected by the studied selection approaches on BIG-bench when using GPT-3.5.}
    \label{fig:impactsize}
\end{figure}

In this section, we further evaluate the impact of sample size across the prompt optimization techniques. Specifically, we conduct experiments with sample sizes of 5, 10, 15, 20, and 30 on GPT-3.5 with BIG-bench, as shown in Figure~\ref{fig:impactsize}. Note that we observe the same trend on GPT-4o-mini with LIAR. Overall, the performance improves as the sample size increases from 5 to 20, then stabilizes or slightly declines beyond 20, suggesting that 20 samples may represent the optimal balance between cost and effectiveness for prompt optimization. 
Across all sample sizes, \textbf{\ourtool consistently outperforms other baselines across all prompt optimization techniques}, even with as few as 5 samples. Anchor-Point generally ranks second, reinforcing our earlier finding (Section~\ref{sec:rq1}) that approaches leveraging model performance data (\ourtool and Anchor-Point) tend to outperform those relying solely on semantic or historical performance data. In addition, \ourtool is more stable than other approaches across different sample sizes. See more details in Appendix~\ref{app:impactsamplesize}.

\subsection{Overhead and cost analysis}\label{sec:rq4}


\begin{table}[]
\centering
\caption{The cost analysis of the studied prompt optimization approaches and warm-up stage for Anchor-Point. The actual cost (in USD) is in parentheses.}

\renewcommand{\arraystretch}{1.2} 
\resizebox{0.5\textwidth}{!}{%
\begin{tabular}{lcc}
\hline

                  & \textbf{Prompt Optimization} & \textbf{Preliminary stage for Anchor-Point}
                  \\ 
\hline
\textbf{APO}               & 637 (0.20)  & 6,124 (0.95)      \\ 
\textbf{APE}               & 231 (0.06)  & 2,277 (0.163)       \\ 
\textbf{EVOPROMPT}         & 2,245 (2.75)     &   22,349 (6.18)    \\ 
\textbf{Average}       &   1,037 (1.00)   &  10,250 (2.43)          \\ \hline\end{tabular}
}
\label{table:4}
\end{table}

\textbf{The overhead of \ourtool is less than 1\%, making it comparable to or better than other baselines.}
Table~\ref{table:time} presents the average execution time of three prompt optimization techniques, after applying coreset selection approaches on BIG-bench using GPT-3.5. \ourtool has a comparable overhead to other baselines. The overhead of \ourtool primarily comes from stage 2, which iteratively identifies and replaces redundant samples based on model performance (2.83 seconds on average). Stage 1 has a similar overhead as Boundary and Clustering (0.45 seconds on average). Anchor-Point has the highest overhead, which requires an additional preliminary stage to evaluate training samples on prompts, resulting in a significantly higher overhead of 51\% and an execution time of approximately 200 seconds. Note that we observe a similar overhead when evaluated on LIAR using GPT-4o-mini.

For Clustering, Boundary, Prediction-based, and \ourtool, since the cost for those approaches only comes from LLM's inference for prompt optimization. For the approach itself, no additional cost is required. The calculation related to clustering and prediction can be done locally by deploying small models. The cost for prompt optimization is presented in Table~\ref{table:4}. Similar to overhead, Anchor-Point requires evaluating the entire model performance during the preliminary stage, therefore additional cost is required. 

\section{Conclusion}\label{sec:conclusion}
We introduced \ourtool, a two-stage approach that enhances coreset selection for prompt optimization by leveraging both semantic and model performance information. Our method first selects representative and diverse samples based on semantic clustering and boundary analysis, followed by an iterative refinement process that integrates real-time model performance information to replace redundant samples with more informative ones. Evaluation on the BIG-bench dataset demonstrated that \ourtool consistently outperforms existing baselines with minimal computational overhead of less than 1\%. Furthermore, the real-time performance-guided refinement approach in \ourtool is universally applicable to other coreset selection methods, enhancing their overall effectiveness and stability.

\section{Limitations}
One limitation is our findings may not generalize to other large language models with different architectures, training data, or capabilities. We encourage future research to evaluate our approach using a diverse set of base models to assess its applicability across different LLMs. We selected three prompt optimization techniques, from different families. We encourage future research to evaluate on more optimization techniques.





\bibliography{main}

\appendix
\newpage
\section{Appendix}
\label{sec:appendix}

\subsection{Experimental setup}
The temperature is set to 0 for inference. All experiments are done in Python 3.10. All experiments were conducted on a machine equipped with a GPU of 24GB, a 24-core CPU, and 24 GB of RAM.

\subsection{Detailed implementation of baselines}\label{sec:baselineImpl}
\textbf{Anchor-Point~\citep{vivek2023anchor}.} To adopt this approach in the context of prompt optimization, we first collect the model's confidence scores on the examples. Specifically, we run the training dataset through an initial set of prompts and then cluster the examples based on their confidence scores across these prompts, following~\citep{vivek2023anchor}. During this stage, we generate 10 prompts using the prompt optimization technique as the initial set. Evaluating confidence on more prompts typically can lead to better clustering results, while inference is expensive. To balance the quality and cost, we select 10 prompts. Also, evaluating the confidence of the entire training dataset is infeasible and expensive. To expedite the process and save inference costs, we first select 200 examples from the entire training data using the same approach of stage 1, and then apply Anchor-Point to select the final evaluation data. \textbf{Prediction-based~\citep{pacchiardi2024100}.} This approach predicts the performance of an instance on an LLM by training a generic assessor based on the performance of each sample in the training set on existing LLMs. We adapt the generic assessor using our dataset, BIG-bench, on GPT-3.5, following the approach outlined by~\cite{pacchiardi2024100}. We then use the trained assessor to predict the performance of each example in the training dataset on a set of initial prompts and subsequently cluster those examples based on their predicted performance, similar to the Anchor-Point.

\subsection{Dataset statistics}\label{app:dataset}
The sizes of the training and testing datasets used in our experiments are presented in Table~\ref{table:5}.
\begin{table}[]
\centering
\caption{The size of Training dataset and testing in our experiments}

\renewcommand{\arraystretch}{1.2} 
\resizebox{0.5\textwidth}{!}{%
\begin{tabular}{lcc}
\hline

                  & \textbf{Training Dataset} & \textbf{Testing dataset}
                  \\ 
\hline
\textbf{Navigation}               & 800  & 200     \\ 
\textbf{Implicatures}                & 392  & 100        \\ 
\textbf{Metaphor Understanding}          & 544  & 136    \\ 
\textbf{Sports Understanding}        & 788  & 198           \\ 
\textbf{Natural Language Inference}        & 588  & 147           \\ 
\textbf{Fake News Detection}        &10240   &1267  \\
\hline\end{tabular}
}
\label{table:5}
\end{table}

\subsection{Ablation analysis on boundary cases selection}

In Stage 1: Diverse sample selection, we select boundary cases to diversify our samples. To understand the impact of the boundary case, we conducted an ablation analysis on the boundary case and the results are shown in Table~\ref{tab:boundaycasesAbl}. Note that we extended our evaluation with one more data LIAR and one more model GPT-4o mini, as suggested by the reviewer. As the results show, boundary case selection makes significant contribution to IPOMP.

\begin{table*}
\centering
\caption{Comparison of \ourtool and \ourtool without boundary cases.}
\label{tab:boundaycasesAbl}
\scriptsize
\renewcommand{\arraystretch}{1.1} 
\resizebox{2.1\columnwidth}{!}{
\begin{tabular}{lcccccccc}
\toprule
 & \multicolumn{2}{c}{\textbf{GPT-4o-mini- Big-Bench}} & \multicolumn{2}{c}{\textbf{GPT-4o-mini-LIAR}} & \multicolumn{2}{c}{\textbf{GPT-3.5-LIAR}} & \multicolumn{2}{c}{\textbf{GPT-3.5-Big-Bench}} \\
\cmidrule(lr){2-3} \cmidrule(lr){4-5} \cmidrule(lr){6-7} \cmidrule(lr){8-9}
 & \textbf{IPOMP} & \textbf{w/o Boundary} & \textbf{IPOMP} & \textbf{w/o Boundary} & \textbf{IPOMP} & \textbf{w/o Boundary} & \textbf{IPOMP} & \textbf{w/o Boundary} \\
\midrule
\textbf{EVOPROMPT} & \textbf{0.758$\pm$0.011} & 0.732$\pm$0.026 & \textbf{0.838$\pm$0.011} & 0.834$\pm$0.035 & \textbf{0.818$\pm$0.015} & 0.785$\pm$0.029 & \textbf{0.776$\pm$0.017} & 0.759$\pm$0.042 \\
\textbf{APO} & \textbf{0.780$\pm$0.012} & 0.738$\pm$0.021 & \textbf{0.836$\pm$0.011} & 0.811$\pm$0.027 & \textbf{0.812$\pm$0.014} & 0.746$\pm$0.025 & \textbf{0.753$\pm$0.009} & 0.723$\pm$0.039 \\
\textbf{APE} & \textbf{0.794$\pm$0.012} & 0.749$\pm$0.026 & \textbf{0.827$\pm$0.013} & 0.790$\pm$0.037 & \textbf{0.832$\pm$0.011} & 0.766$\pm$0.027 & \textbf{0.742$\pm$0.010} & 0.715$\pm$0.031 \\
\textbf{Average} & \textbf{0.778$\pm$0.011} & 0.740$\pm$0.024 & \textbf{0.833$\pm$0.012} & 0.812$\pm$0.033 & \textbf{0.820$\pm$0.013} & 0.765$\pm$0.027 & \textbf{0.757$\pm$0.012} & 0.732$\pm$0.037 \\
\bottomrule
\end{tabular}
}
\end{table*}

\subsection{Results of impact of hyper-parameters}\label{sec:rq5}

\begin{figure}
    \centering
    \includegraphics[width=0.5\textwidth]{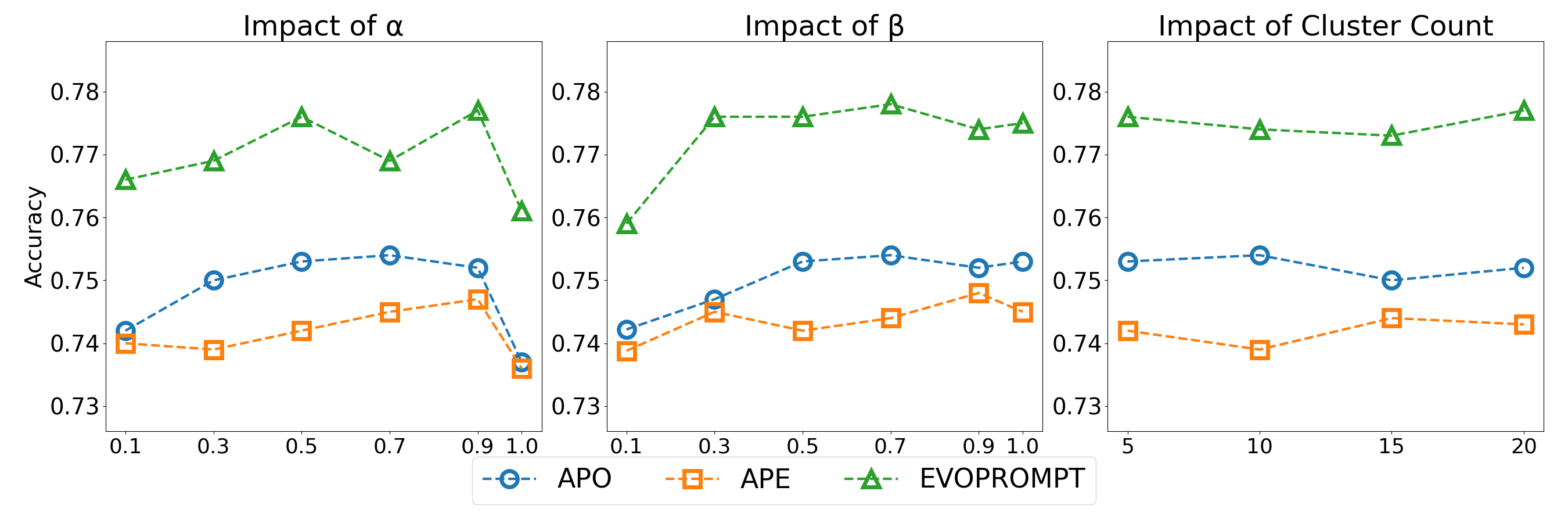}
    \caption{The impact of different values of hyper-parameters $\alpha$ ,  $\beta$  and cluster size $K$ on the effectiveness of APE, APO, and EVOPROMPT.}
    \label{fig:impactOfHyper}
\end{figure}

\begin{table*}[]
\centering
\caption{Comparison of the effectiveness (in terms of accuracy) and stability (in terms of deviation) of the studied prompt optimization approaches with different evaluation data sampling approaches on BIG-bench when using GPT-3.5.}\label{tab:impactofsize}
\scriptsize
\renewcommand{\arraystretch}{1.1}
\resizebox{0.95\textwidth}{!}{%
\begin{tabular}{lccccccc}
\hline
 & \textbf{Size} & \textbf{Random} & \textbf{Boundary} & \textbf{Clustering} & \textbf{Anchor-Point} & \textbf{Prediction-based} & \textbf{\ourtool} \\ 

\hline
\multirow{5}{*}{\textbf{APO}}& 5 & 0.678$\pm$0.013 & 0.672$\pm$0.043    &0.674$\pm$0.021   & \textbf{0.685\textcolor{red}{$\uparrow$}$\pm$0.029} &0.675$\pm$0.039   & 0.683$\pm$0.017\textcolor{green}{$\downarrow$}  \\ 
& 10 & 0.702$\pm$0.031 &0.714$\pm$0.029   &0.708$\pm$0.019 &0.720$\pm$0.015 &0.704$\pm$0.032    & \textbf{0.722\textcolor{red}{$\uparrow$}$\pm$0.010\textcolor{green}{$\downarrow$}} \\ 
& 15  & 0.703$\pm$0.023 & 0.718$\pm$0.042   &0.720$\pm$0.022    & 0.744$\pm$0.020 & 0.732$\pm$0.038    & \textbf{0.750\textcolor{red}{$\uparrow$}$\pm$0.008\textcolor{green}{$\downarrow$}} \\ 
& 20 & 0.691$\pm$0.040 & 0.718$\pm$0.052   & 0.723$\pm$0.025   & 0.750$\pm$0.020 & 0.743$\pm$0.042    & \textbf{0.753\textcolor{red}{$\uparrow$}$\pm$0.009\textcolor{green}{$\downarrow$}} \\ 
& 30 &0.692$\pm$0.023 &0.714$\pm$0.038    &0.720$\pm$0.028      &0.748$\pm$0.028 &0.740$\pm$0.038    & \textbf{0.752\textcolor{red}{$\uparrow$}$\pm$0.011\textcolor{green}{$\downarrow$}} \\ \hline

\multirow{5}{*}{\textbf{APE}} & 5 & 0.693$\pm$0.022 &0.677$\pm$0.024   & 0.688$\pm$0.023 &0.702$\pm$0.021  & 0.698$\pm$0.029 & \textbf{0.708\textcolor{red}{$\uparrow$}$\pm$0.019\textcolor{green}{$\downarrow$}} \\ 
& 10 & 0.711$\pm$0.026 &0.712$\pm$0.024   &0.712$\pm$0.020   &0.722$\pm$0.026  &0.712$\pm$0.030  & \textbf{0.724\textcolor{red}{$\uparrow$}$\pm$0.014\textcolor{green}{$\downarrow$}} \\ 
& 15 &0.727$\pm$0.041  &  0.708$\pm$0.042   &0.702$\pm$0.032 &0.728$\pm$0.032  &0.690$\pm$0.042     & \textbf{0.736\textcolor{red}{$\uparrow$}$\pm$0.013\textcolor{green}{$\downarrow$}} \\ 
& 20  & 0.722$\pm$0.037 & 0.708$\pm$0.045   & 0.684$\pm$0.032   & 0.727$\pm$0.035 & 0.707$\pm$0.048    & \textbf{0.742\textcolor{red}{$\uparrow$}$\pm$0.010\textcolor{green}{$\downarrow$}} \\ 
& 30  &0.719$\pm$0.035   &0.710$\pm$0.040  &0.698$\pm$0.034    &0.727$\pm$0.033   &  0.710$\pm$0.032   & \textbf{0.740\textcolor{red}{$\uparrow$}$\pm$0.012\textcolor{green}{$\downarrow$}}\\ \hline

\multirow{5}{*}{\textbf{EVOPROMPT}} & 5 &0.704 $\pm$0.034     &0.696$\pm$0.027 &0.708$\pm$0.018   &0.703$\pm$0.027  &0.673$\pm$0.034 & \textbf{0.712\textcolor{red}{$\uparrow$}$\pm$0.020\textcolor{green}{$\downarrow$}} \\ 
& 10 &0.729$\pm$0.029  &0.730$\pm$0.026   &\textbf{0.740\textcolor{red}{$\uparrow$}$\pm$0.032}    &0.736$\pm$0.018  &0.712$\pm$0.020    & 0.732$\pm$0.017\textcolor{green}{$\downarrow$} \\ 
& 15 &0.739$\pm$0.019 &0.738$\pm$0.029   &0.742$\pm$0.019   &0.750$\pm$0.019   & 0.739$\pm$0.029   & \textbf{0.765\textcolor{red}{$\uparrow$}$\pm$0.011\textcolor{green}{$\downarrow$}}\\ 
& 20 & 0.743$\pm$0.028 & 0.757$\pm$0.022   & 0.759$\pm$0.031   & 0.774$\pm$0.028 & 0.758$\pm$0.025    & \textbf{0.776\textcolor{red}{$\uparrow$}$\pm$0.017\textcolor{green}{$\downarrow$}}  \\ 
& 30 &0.747$\pm$0.032  &0.754$\pm$0.032    &0.759$\pm$0.034    & \textbf{0.780\textcolor{red}{$\uparrow$}$\pm$0.036} &  0.755$\pm$0.030   & 0.778$\pm$0.018\textcolor{green}{$\downarrow$} \\ \hline
\end{tabular}
}
\end{table*}

We examine the impact of key hyper-parameters in \ourtool: $\alpha$ which controls the proportion of samples selected based on semantic clustering in stage 1, and $K$ which controls the cluster size in stage 1, and $\beta$ which determines the proportion of redundant samples to be replaced in stage 2. Figure~\ref{fig:impactOfHyper} presents the results of BIG-bench when using GPT-3.5. 

As $\alpha$ increases from 0.1 to 0.9, the accuracy of \ourtool consistently improves across all three prompt optimization techniques. However, when $\alpha$ exceeds 0.9, the performance of APE, APE, and EVOPROMPT degrades significantly. This observation suggests that while incorporating a small portion of boundary cases can enhance the diversity of the evaluation data and improve the performance of \ourtool, relying too heavily on boundary cases can negatively impact the overall effectiveness (i.e., $\alpha <$ 0.9). In contrast, selecting samples purely based on semantic clustering results in suboptimal performance (i.e., $\alpha =$ 1).


In terms of $\beta$, as it increases from 0.1 to 0.7, the performance of the three prompt optimization techniques improves gradually, particularly noticeable in the range from 0.1 to 0.3. For APO and EVOPROMPT, their performance still gets improved until $\beta$ reaches 0.7 and gets degraded after 0.7. For APE, its performance degrades from 0.3 to 0.5, and improves after 0.5. This enhancement is attributed to the replacement of redundant examples during each iteration, which effectively increases the diversity of the evaluation samples based on the feedback from the model in real-time.

We also investigate the impact of the cluster size $K$. As Figure~\ref{fig:impactOfHyper} shows, the size of clusters does not impact the effectiveness of \ourtool significantly.

\begin{figure*}[ht]
    \centering
    \includegraphics[width=\textwidth]{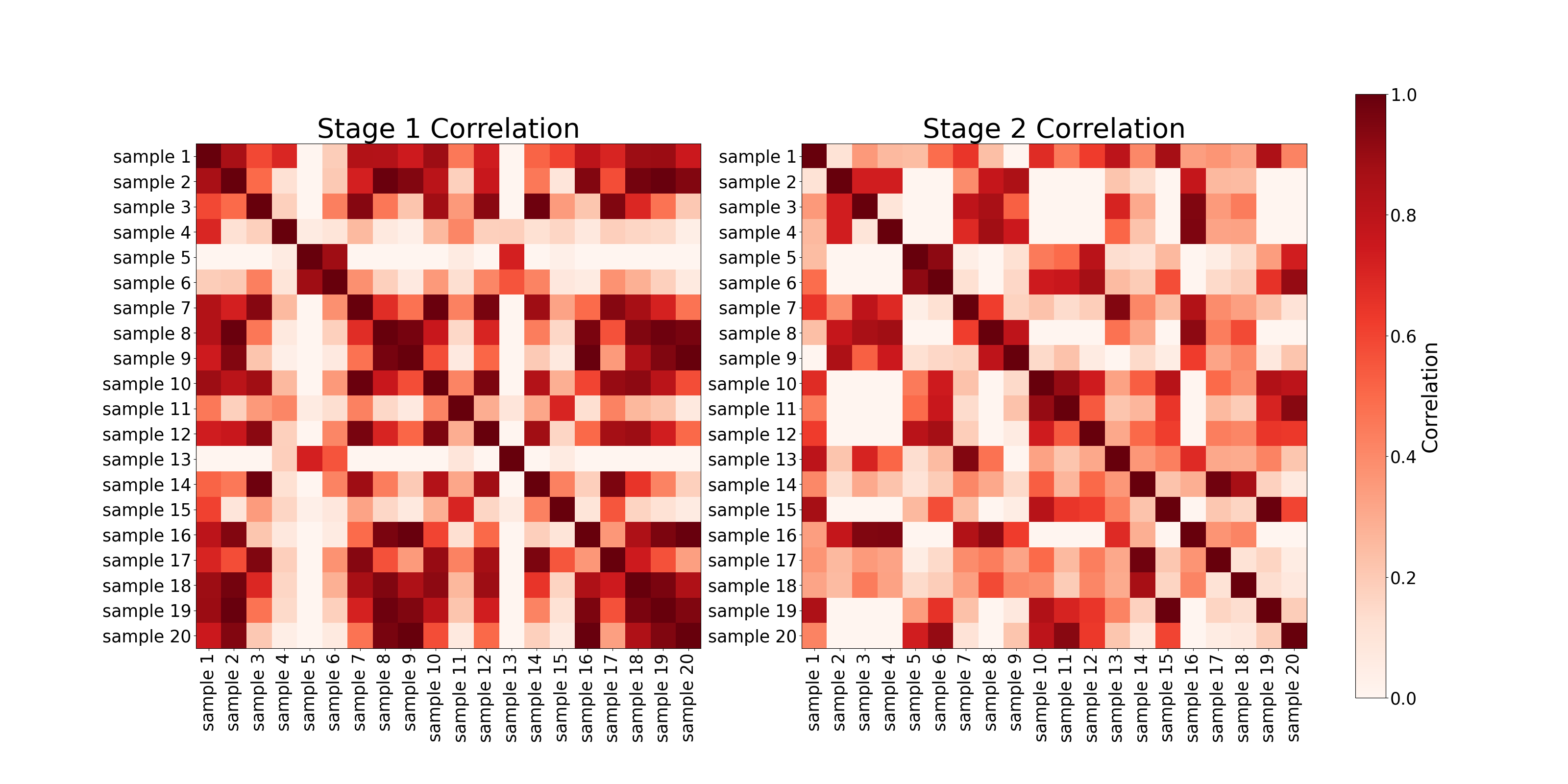} 
    \caption{Correlation among samples before and after the first round of real-time model performance-guided refinement in APO on the Implicatures dataset. Each cell represents the correlation between a pair of samples.}
    \label{fig:step12} 
\end{figure*}

\subsection{Detailed results of impact of sample size}\label{app:impactsamplesize}

The accuracy and standard deviation of prompt optimization techniques over different sizes of samples selected by the studied approaches are presented in Table~\ref{tab:impactofsize}.

\subsection{Impact of replacement strategy in Stage 2}. 
To verify our hypothesis that selecting dissimilar examples yields significantly lower performance correlation compared to both random and similar selection, we conducted an ablation study to compare three sample replacement strategies:
\begin{itemize}
    \item \textbf{Dissimilar}: Replacing redundant samples with the most dissimilar samples from the training dataset (our proposed method).
    \item \textbf{Random}: Replacing redundant samples with randomly selected samples from the training dataset.
    \item \textbf{Similar}: Replacing redundant samples with the most similar samples from the training dataset.
\end{itemize}

We measured the correlation between the performance of the original samples and replaced samples using the above replacement strategies. The results shown in Table~\ref{tab:correlation_summary} confirm our hypothesis. This validates the importance of our replacement strategy in influencing model behavior.

\begin{table}[]
\centering
\scriptsize
\caption{Comparison of five-number summary of correlation among the samples selected using different strategies.}
\label{tab:correlation_summary}
\begin{tabular}{lccccc}
\toprule
 & Min & Q1 & Median & Q3 & Max \\
\midrule
Similar & 0.623 & 0.938 & 0.960 & 0.978 & 0.999 \\
Random & 0.002 & 0.418 & 0.634 & 0.882 & 0.943 \\
Dissimilar & 0.002 & 0.109 & 0.400 & 0.480 & 0.681 \\
\bottomrule
\end{tabular}
\end{table}

\subsection{Case study}\label{sec:casestudy}

Figure~\ref{fig:step12} presents the correlation among samples before and after applying real-time model performance-guided refinement in APO. As we can observe, after the refinement, the redundancy of examples (correlation $> 0.9$) are significantly reduced from 19\% to 10\%.

We present the selected examples in stage 1 and after the first round of refinement in stage 2 below.

\clearpage
\begin{tcolorbox}[
    boxrule=0.1mm, boxsep=1mm,
    title=Selected Examples by stage 1 of \ourtool,
    colback=white, colframe=black!75!black, 
    width=\textwidth, center title, enhanced,
    halign=center
]
\footnotesize
\centering
\begin{itemize}
    \setlength{\itemsep}{0.5pt}  
    \setlength{\parsep}{0pt} 

    \item Example 1: Speaker 1: \textit{'This is a costume?'} Speaker 2: \textit{'Aaaiyyyy... worked on it all night long!'}
    \item Example 2: Speaker 1: \textit{'Do you love me?'} Speaker 2: \textit{'My love for you is as deep as the ocean.'}
    \item Example 3: Speaker 1: \textit{'Did you sleep well last night?'} Speaker 2: \textit{'Last night, I slept like a log.'}
    \item Example 4: Speaker 1: \textit{'Do you think that Dr. Luby will organize a theatre trip to New York this year?'} Speaker 2: \textit{'I have already signed up for it.'}
    \item Example 5: Speaker 1: \textit{'Did you report Private Barnes to your superiors?'} Speaker 2: \textit{'I remember thinking very highly of Private Barnes, and not wanting to see his record tarnished by a formal charge.'}
    \item Example 6: Speaker 1: \textit{'I bought the wrong math book. Here is the receipt. Can I get my money back?'} Speaker 2: \textit{'Not after ten days. But you can exchange something for it.'}
    \item Example 7: Speaker 1: \textit{'Does it bother you that your husband goes away on long business trips?'} Speaker 2: \textit{'Absence makes the heart grow fonder.'}
    \item Example 8: Speaker 1: \textit{'Did you order the code red?'} Speaker 2: \textit{'You're goddamn right.'}
    \item Example 9: Speaker 1: \textit{'My client is taking me to a really fancy restaurant tonight. So I am wearing this new cologne. I got a sample of it from a magazine. Can you smell it?'} Speaker 2: \textit{'From across the room. But it is not exactly subtle. Is it?'}
    \item Example 10: Speaker 1: \textit{'Are you a Dodgers fan?'} Speaker 2: \textit{'I don't like baseball.'}
    \item Example 11: Speaker 1: \textit{'Is everyone comfortable?'} Speaker 2: \textit{'Everyone is on pins and needles.'}
    \item Example 12: Speaker 1: \textit{'I feel horrible. Debbie was furious that I lost her notes. Do you think I should apologize to her again?'} Speaker 2: \textit{'If I were you, I would cool off for some days before I talk to her again.'}
    \item Example 13: Speaker 1: \textit{'Should I decide now?'} Speaker 2: \textit{'Why don't you go home and sleep on it?'}
    \item Example 14: Speaker 1: \textit{'Did I do it well?'} Speaker 2: \textit{'You were as brave as a lion.'}
    \item Example 15: Speaker 1: \textit{'Are you coming with me to the exhibition today?'} Speaker 2: \textit{'I made plans with Susan to go to the exhibition tomorrow afternoon.'}
    \item Example 16: Speaker 1: \textit{'Does it rain here nowadays?'} Speaker 2: \textit{'It's been raining for 40 days and 40 nights.'}
    \item Example 17: Speaker 1: \textit{'Are you planning to buy a house?'} Speaker 2: \textit{'I really want a place to call my own.'}
    \item Example 18: Speaker 1: \textit{'Is it a good product?'} Speaker 2: \textit{'They had put a lot of thought into making it.'}
    \item Example 19: Speaker 1: \textit{'Is that book about lullabies?'} Speaker 2: \textit{'It is about symphonies.'}
    \item Example 20: Speaker 1: \textit{'But aren't you afraid?'} Speaker 2: \textit{'Ma'am, sharks never attack anybody.'}
\end{itemize}

\end{tcolorbox}

\clearpage

\begin{tcolorbox}[
    enlarge left by=-0.1cm, enlarge right by=-1cm, 
    boxrule=0.1mm, boxsep=1mm,
    title=Selected Examples after first round of refinment at stage 2.,
    colback=white, colframe=black!75!black, 
    width=\textwidth, center title, enhanced,
    halign=center
]
\footnotesize
\centering
Here are selected examples from Stage 2 of our tool's processing:
\begin{itemize}
    \setlength{\itemsep}{0.5pt}  
    \setlength{\parsep}{0pt} 

    \item Example 1: Speaker 1: \textit{'This is a costume?'} Speaker 2: \textit{'Aaaiyyyy... worked on it all night long!'}
    \item Example 2: Speaker 1: \textit{'Do you love me?'} Speaker 2: \textit{'My love for you is as deep as the ocean.'}
    \item Example 3: Speaker 1: \textit{'Did you sleep well last night?'} Speaker 2: \textit{'Last night, I slept like a log.'}
    \item Example 4: Speaker 1: \textit{'My client is taking me to a really fancy restaurant tonight. So I am wearing this new cologne. I got a sample of it from a magazine. Can you smell it?'} Speaker 2: \textit{'From across the room. But it is not exactly subtle. Is it?'}
    \item Example 5: Speaker 1: \textit{'I bought the wrong math book. Here is the receipt. Can I get my money back?'} Speaker 2: \textit{'Not after ten days. But you can exchange something for it.'}
    \item Example 6: Speaker 1: \textit{'I feel horrible. Debbie was furious that I lost her notes. Do you think I should apologize to her again?'} Speaker 2: \textit{'If I were you, I would cool off for some days before I talk to her again.'}
    \item Example 7: Speaker 1: \textit{'Should I decide now?'} Speaker 2: \textit{'Why don't you go home and sleep on it?'}
    \item Example 8: Speaker 1: \textit{'Are you planning to buy a house?'} Speaker 2: \textit{'I really want a place to call my own.'}
    \item Example 9: Speaker 1: \textit{'Is that book about lullabies?'} Speaker 2: \textit{'It is about symphonies.'}
    \item Example 10: Speaker 1: \textit{'Are you angry at me?'} Speaker 2: \textit{'To err is human, to forgive divine.'}
    \item Example 11: Speaker 1: \textit{'Does it rain here nowadays?'} Speaker 2: \textit{'It's been raining for 40 days and 40 nights.'}
    \item Example 12: Speaker 1: \textit{'That cake looks delicious. Aren't you going to have some with me?'} Speaker 2: \textit{'I am watching my calorie intake.'}
    \item Example 13: Speaker 1: \textit{'Does she know how to play the piano?'} Speaker 2: \textit{'Now it is like second nature to her.'}
    \item Example 14: Speaker 1: \textit{'Do you have these journals?'} Speaker 2: \textit{'I can have them flown here from Geneva in an hour.'}
    \item Example 15: Speaker 1: \textit{'Do you have any agricultural background?'} Speaker 2: \textit{'I used to work in an office.'}
    \item Example 16: Speaker 1: \textit{'Do you have field hands that help you?'} Speaker 2: \textit{'We work the land alone.'}
    \item Example 17: Speaker 1: \textit{'I feel horrible. Debbie was furious that I lost her notes. Do you think I should apologize to her again?'} Speaker 2: \textit{'If I were you, I would cool off for some days before I talk to her again.'}
    \item Example 18: Speaker 1: \textit{'You have it, then?'} Speaker 2: \textit{'I had to slit a few throats to get it.'}
    \item Example 19: Speaker 1: \textit{'Were you hiding from me?'} Speaker 2: \textit{'I didn't want to scare you.'}
    \item Example 20: Speaker 1: \textit{'Are you coming with me to the exhibition today?'} Speaker 2: \textit{'I made plans with Susan to go to the exhibition tomorrow afternoon.'}
\end{itemize}

\end{tcolorbox}


\end{document}